\documentclass[conference]{IEEEtran}
\IEEEoverridecommandlockouts
\usepackage{cite}
\usepackage{amsmath,amssymb,amsfonts}
\usepackage{indentfirst}
\usepackage{algorithm} 
\usepackage{algpseudocode}
\usepackage{graphicx}
\usepackage{amsmath}
\usepackage{float}
\usepackage{textcomp}
\usepackage{xcolor}
\usepackage{multirow}
\usepackage[hidelinks]{hyperref}
\def\BibTeX{{\rm B\kern-.05em{\sc i\kern-.025em b}\kern-.08em
    T\kern-.1667em\lower.7ex\hbox{E}\kern-.125emX}}
\usepackage[left=1.01in,right=1.01in,top=1.9cm]{geometry}

\begin{document}

\title{Real-time Stress Detection on Social Network Posts using Big Data Technology}

\author{
  \IEEEauthorblockN{Hai-Yen Phan Nguyen. A\textsuperscript{*,\textdagger},
  Phi-Lan Ly\textsuperscript{*,\textdagger},
  Duc-Manh Le\textsuperscript{*,\textdagger},
  Trong-Hop Do \textsuperscript{*,\textdagger} }
  \IEEEauthorblockA{\textsuperscript{*} University of Information Technology \\ \textsuperscript{\textdagger}Vietnam National University Ho Chi Minh City,Vietnam}
  \{21521698, 21520319, 21521116\}@gm.uit.edu.vn,\\
  Corresponding author: hopdt@uit.edu.vn
}\maketitle


\begin{abstract}
In the context of modern life, especially in the industry 4.0 on the online space, emotions and moods of people are often expressed through posts on social media platforms. The trend of sharing stories, thoughts and feelings of people on social media platforms creates a huge and potential data source for the field of Big Data. This poses a challenge and at the same time creates an opportunity for research and application of technology to develop more automatic and accurate methods in detecting stress of social media users. In this work, we built a real-time system for detecting stress from online posts based on the dataset “Dreaddit: A Reddit Dataset for Stress Analysis in Social Media”, collected from 187,444 posts in five different domains on Reddit. Each domain includes texts containing stressful or non-stressful content as well as different ways of expressing stress in the post, of which there are 3,553 labeled lines to form the training dataset. We used Apache Kafka, PySpark combined with AirFlow to build the model and deploy the system. LogisticRegression gives the highest results for new streaming data, reaching 69,39\% for measuring accuracy and 68,97 for measuring F1-scores.
\end{abstract}
\section{Introduction}
With the rapid development of information technology, social networking sites have become an indispensable part of the lives of many people today. According to a report by Statista, as of October 2023, more than 60\% of the world’s population uses social media. The large number of users leads to a huge number of posts being shared every day, along with the modern life with many pressures from work, study, family, society, etc, making stress a common and easily observable condition than ever before.\\

Within the scope of this topic, stress can be understood as a state of anxiety or mental pressure that can arise from various causes, such as work pressure, study, life, relationships, etc. Stress can not only be recognized by biological parameters but it can also be detected by emotions, questions and ways of posing problems through posts on social networking sites, thereby further reinforcing the feasibility of stress detection. Prolonged or excessive stress can affect the quality of life, work and study performance, as well as cause physical and mental illnesses.\\

The detection of stress has been widely discussed in numerous research works due to its significant contributions to society and human life. Stress can be identified physically through biological parameters such as heart rate (HR), respiration, electromyography (EMG), galvanic skin response (GSR), as studied by Ghaderi et al. (2015)\textbf{\textcolor{red}{~\cite{ghaderi2015}}}. Additionally, David Liu and colleagues (2015)\textbf{\textcolor{red}{~\cite{salai2016}}} proposed predicting stress levels solely from Electrocardiogram (ECG), or S. Sriramprakash and team applied machine learning techniques like Support Vector Machine (SVM) and K-Nearest Neighbors (KNN)\textbf{\textcolor{red}{~\cite{sriramprakash2017}}} to detect stress using ECG and GSR during work. These studies collected human physical reactions to stress, but a drawback is the need for direct and specialized data collection tools from subjects, making it challenging to meet the demands of current society.\\

With the proliferation of stress on social media, we believe that stress can be observed and studied entirely from massive text data related to stress. Research on stress detection on social media platforms, such as the study by Winata et al. (2018)\textbf{\textcolor{red}{~\cite{salai2016}}}, utilized Long Short-Term Memory (LSTM) networks to detect stress in speech and Twitter data. More recently, Arfan Ahmed and colleagues (2022)\textbf{\textcolor{red}{~\cite{ahmed2022}}} investigated data from the Facebook platform, while Lin et al. (2017)\textbf{\textcolor{red}{~\cite{kim2014}}} detected stress in short blog posts using Convolutional Neural Networks (CNN).

In this work, we focused on building a system that can receive information as posts on the social networking site Reddit and automatically detect the stress level of the post. From there, it serves as a basis for social networking sites to provide timely support to users.

The uniqueness of our work lies in using data from the Reddit social media platform. This data is typically longer than datasets from other platforms, allowing for a clearer analysis of factors in the data and posing challenges in incorporating data into machine learning models. Furthermore, the Reddit data possesses special attributes not found in other platforms, providing additional elements for analysis and enhancing the comprehensiveness of the machine learning model.
\section {Dataset}
\subsection{About dataset}
\begin{figure}[h]
\includegraphics[width=\linewidth]{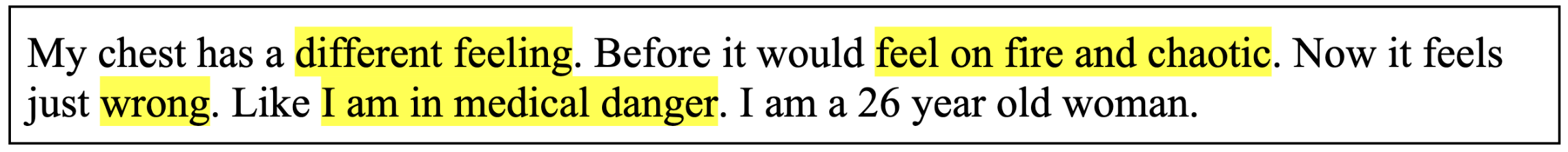}
\caption{Illustration of a stressed post in the Dreaddit dataset} \label{fig1}
\end{figure}

We utilize the Dreaddit\textbf{\textcolor{red}{~\cite{turcan2019}}} dataset, which comprises posts from the social media platform Reddit - a place where users can share, discuss, and explore various topics organized into different subreddits. These posts are categorized into five classes, each representing a specific field, including texts with stress-related characteristics or without stress, expressed in various ways. Specifically, the categories are as follows:\\
\\
- Interpersonal Conflict: Belonging to the abuse and social categories.\\
- Mental Illness: Belonging to the anxiety and post-traumatic stress disorder (PTSD) categories.\\
- Financial Need: Belonging to the financial category.\\
\\
The dataset used contains 3,553 labeled posts, and the language aspect is analyzed using Linguistic Inquiry and Word Count (LIWC).\\
\\
Figure\textbf{\textcolor{red}{~\ref{fig1}}} provides an illustrative example of a post exhibiting signs of stress, with highlighted phrases indicating signs that the author is experiencing stress.
\subsection{Dataset Analysis}
\begin{figure}[h]
\centering
\includegraphics[width=\linewidth]{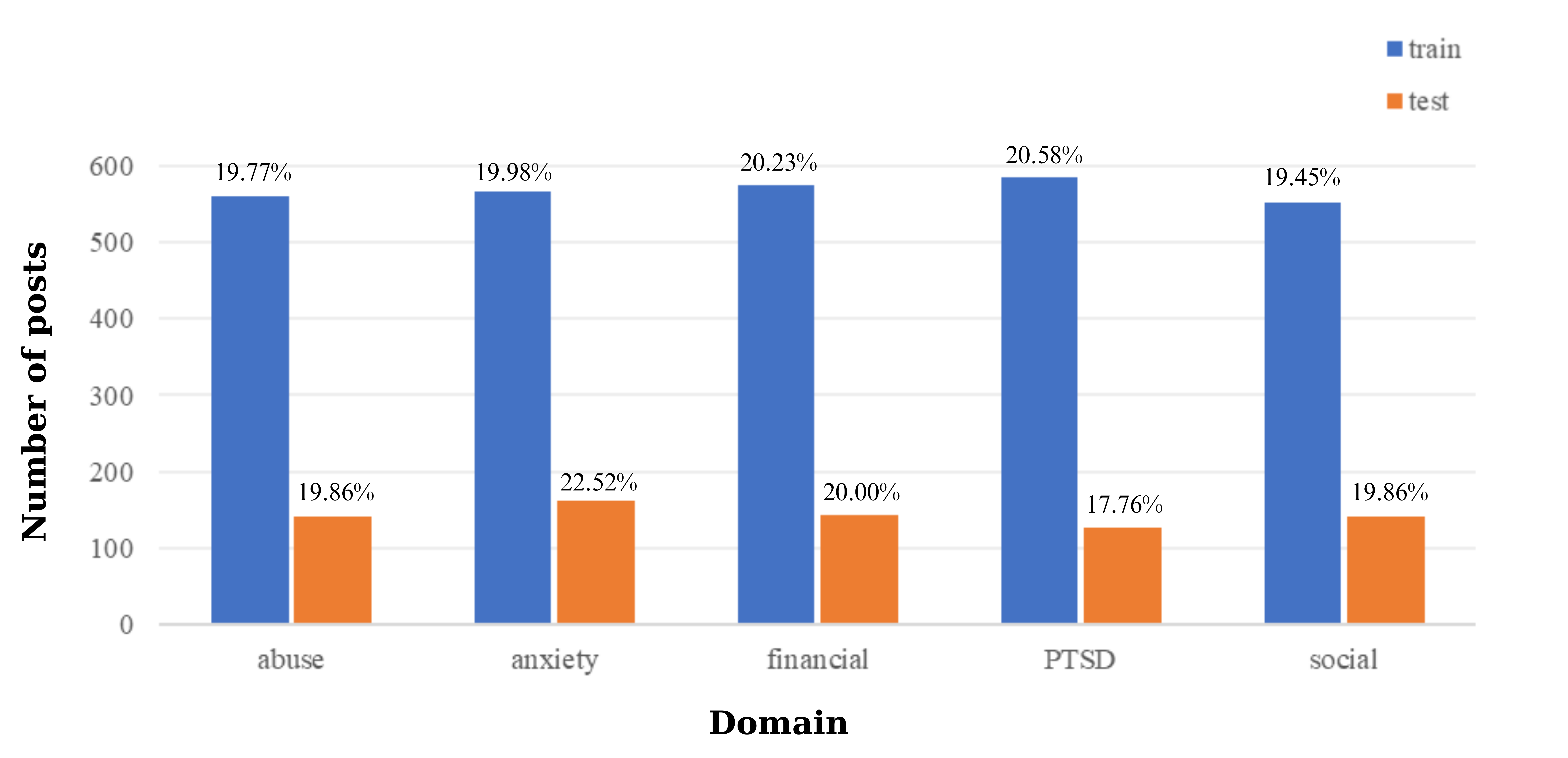}
\caption{\centering The chart of representation of the number and proportion of posts in 5 domains on the training and test set} \label{fig2}
\end{figure}

Figure\textbf{\textcolor{red}{~\ref{fig2}}} shows the distribution and ratio of posts in 5 domains in the train/test set of the Dreaddit dataset that we use.\\

The distribution of posts on the train and test sets does not differ too much, only fluctuating by no more than 1\% compared to the 20\% level on the total number of posts, ensuring that the machine learning model trained on the train set can generalize well when applied to new data, that is, the test set. This means that the model will not be biased too much towards some specific domains when evaluated on the test set. This uniformity also helps to ensure that the model does not learn the distinctive features that only appear in some specific domains that cannot be generalized. In addition, posts labeled as 1 (stress) account for 52.27\%, which is a positive sign that the dataset being used is quite balanced. Balancing data helps to ensure that the machine learning model will be trained effectively on both classes or groups that are less common.\\

\begin{figure}[h]
\centering
\includegraphics[width=\linewidth]{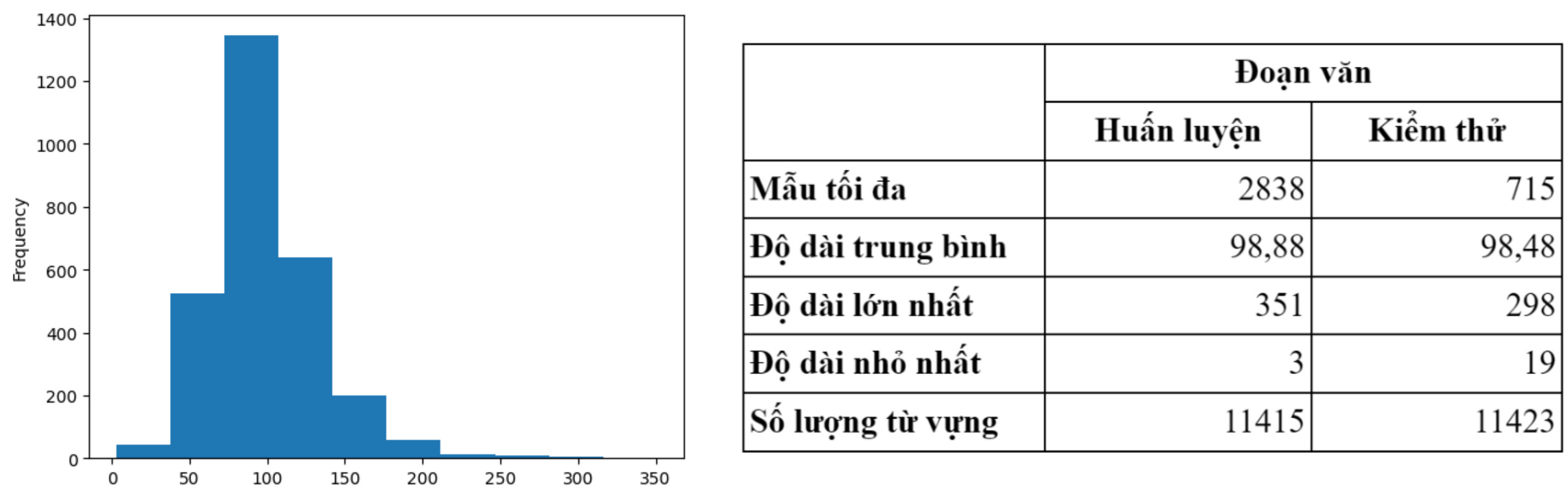}
\caption{\centering Survey results of the lengths of posts in the dataset} \label{fig3}
\end{figure}

Figure\textbf{\textcolor{red}{~\ref{fig3}}} shows the data analysis of the dataset. We can see that the length of the data samples is mostly around 100 words for both the training and testing sets, with a total of nearly 1350 samples, and it decreases on both sides of the peak. Among them, the longest article in the training set has 351 words and the one in the testing set has 298 words. Because the dataset is not fully processed, there are also some articles with relatively short lengths, the most notable one being the shortest article in the training set with only 3 words. And the total vocabulary size of both sets is also similar, ranging from 11,415 to 11,423 words.
\section {Methodology}
\subsection{Data preprocessing} \label{4.1}
The data preprocessing workflow plays a crucial role in every data analysis system, especially for data collected on social media platforms because its complexity is influenced by many factors. Social media is often considered a valuable data repository, but it comes with text formats that are complex, including various elements like links, special characters, emoticons, teencode, and more. Therefore, before feeding the data into models, it undergoes preprocessing through the following steps: removing noise components, removing stopwords, lowercase converting, and tokenization.
\subsection{Feature Engineering}
Feature engineering is an important step before feeding data into models, ensuring that the input data adheres to a common format and can be compared or utilized efficiently in machine learning models. Feature engineering helps reduce variability and standardizes measurement units, creating favorable conditions for the training process. We have applied the following feature engineering steps: String indexer, word embedding using Word2Vec models, vector assembler, and data standardization.
\subsection{Models}
\subsubsection{Machine Learning models}
\paragraph{LogisticRegression} is a linear model commonly used for classification problems. The foundation of this model is the Sigmoid function, which transforms the linear combination of input features into the probability of falling into a specific class. The training process of Logistic Regression involves optimizing the weights through Gradient Descent to maximize the accuracy of predictions.
\paragraph{RandomForest} is a model built from multiple independent decision trees. Each tree is constructed on a random subset of data and utilizes a random set of features. The prediction result of Random Forests is the outcome of voting from all member trees, creating a robust, stable model with good resistance to overfitting.
\paragraph{Decision Trees} is a greedy algorithm that divides data into groups based on feature values. Each partition is chosen greedily to optimize the purity of the child groups. Decision Trees is often used to enhance the interpretability of important features in the data.
\paragraph{Gradient-Boosted Trees}  combine multiple decision trees to create a powerful model. The training process is sequential, with each new tree built to correct the prediction errors of the previous trees. This results in a model capable of learning from weaknesses and optimizing accuracy on the training dataset. 
\paragraph{Support Vector Machines (SVM)} is a classification and regression model that uses an optimal margin between classes. SVM finds the dividing boundary by minimizing the distance between the boundary and the nearest data points, known as support vectors. SVM performs well in high-dimensional feature spaces and is effective for non-linear data.\\
The decision function of the SVM model is given by:
\begin{align}
f(\mathbf{x}) &= \mathbf{w} \cdot \mathbf{x} + b \label{eq:linear_regression} \\
\text{Where,} \quad & \nonumber \\
\mathbf{x} & \text{ is the feature vector of the input sample,} \nonumber \\
\mathbf{w} & \text{ is the weight vector,} \nonumber \\
b & \text{ is the bias value.} \nonumber
\end{align}
The prediction for a sample x is determined by checking the sign of f(x):
\[
y =
\begin{cases}
    1 & \text{if } f(\mathbf{x}) \geq 0 \\
   -1 & \text{if } f(\mathbf{x}) < 0
\end{cases}
\]
\subsubsection{Deep Learning models}
\paragraph{BERT (Bidirectional Encoder Representations from Transformers)} is an advanced natural language processing model developed by Google AI, renowned for its ability to understand natural language representations. Trained on a large amount of language data, this model is suitable for various applications such as text classification, language translation, and many other tasks.
\paragraph{RoBERTa (Robustly optimized BERT approach)} is an advanced language model, a variant of BERT developed by Facebook AI Research (FAIR). RoBERTa is known for its improved training process and enhanced network architecture performance. This model is commonly used to address issues such as text classification, language detection, and related tasks.
\paragraph{DistilBERT} is a "lightweight" version of BERT, optimized to achieve good performance with lower computational costs. Developed by Hugging Face, this model is suitable for tasks that demand flexibility and fast processing speed, such as text classification and summarization.
\paragraph{XLNet} is a natural language processing model developed by Google AI and Carnegie Mellon University. XLNet's unique feature is its ability to integrate "permutation language modeling," enhancing its understanding of natural language. This model is suitable for tasks that require detailed text comprehension and understanding relationships between words.
\paragraph{Electra Base Discriminator} is an advanced model that utilizes the "discriminative pretraining" method instead of the "masked language modeling" method used by BERT. This approach helps the Electra model achieve good performance with less data. Electra is commonly used in applications such as text classification and summarization.
\subsection{Bigdata and real-time technologies}
\subsubsection{Apache Spark}
is an open-source distributed data processing system that utilizes in-memory processing, ensuring high-speed processing for handling and analyzing large datasets. Apache Spark is divided into two main components: Spark Core and the Library. Users submit jobs to Spark Core, and through the Spark Core API, it processes, executes, and responds to users using programming languages such as Scala, Python, Java, and R.\\
\\
In this project, we leverage Spark SQL and Spark Structured Streaming to process data on the Reddit platform. MLlib is applied to build machine learning models within Spark. To deploy a real-time stress detection system, we use Pyspark (2), which is a Python interface for Apache Spark.
\subsubsection{Apache Kafka}
is a platform developed by LinkedIn, then transferred to Apache Software Foundation and released as an open source. Kafka is a real-time processing system, handling distributed data with high performance and reliability. With a simple and flexible structure, Kafka has become one of the leading technologies in processing critical data streams, supporting the development of event-based applications and integration between different systems.\\

Within the scope of the project, we use Kafka to meet the demand for streaming data in real time, as well as creating a temporary storage for the data being used.\\
\section{System architecture}
\begin{figure}
\centering
\includegraphics[width=\linewidth]{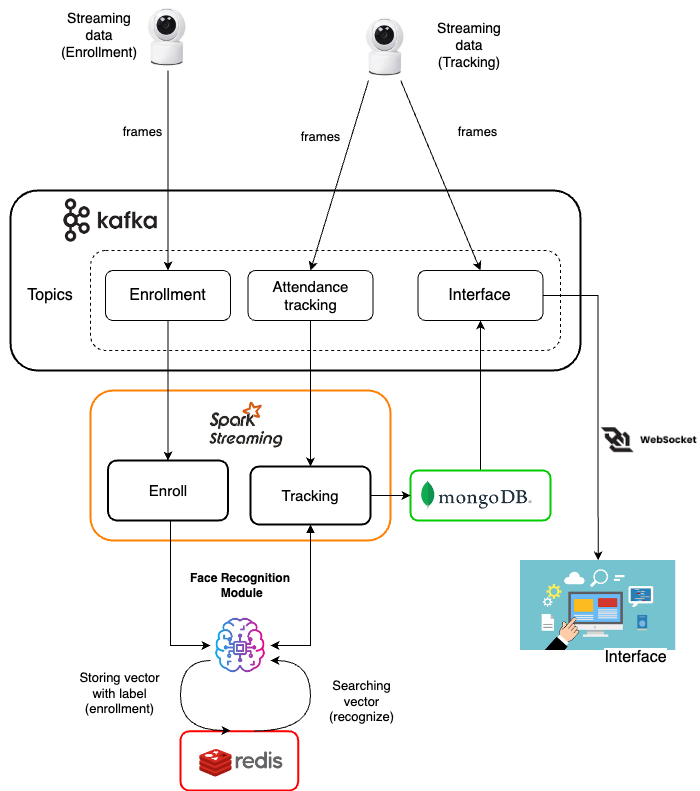}
\caption{Architecture of real-time stress detection system} \label{architecture}
\end{figure}
We built and illustrated a real-time stress detection system through figure\textbf{\textcolor{red}{~\ref{architecture}}}, which consists of two main components: an offline and online processing system, combined with the use of Apache Airflow to manage and schedule tasks.
\subsection{Apache Airflow}
Using Airflow helps ensure that the data preprocessing steps, the model training process and the performance analysis stages are performed automatically. At the same time, Airflow provides a visual interface to monitor the process and confirm success or handle errors when necessary.\\

- In the \textit{offline processing system}, we use Airflow to schedule the workflows in the stages of data preprocessing, model training and model performance evaluation.\\

- In the \textit{online processing system}, Airflow is used to schedule the tasks in the stages of streaming data through Kafka and Spark Structured Streaming, data preprocessing and stress detection in real time.
\subsection{Offline processing system}
The offline processing system was built to train and test the models in order to find the best model to apply in the online stress detection pipeline. We use Machine Learning models in Spark MLlib including: SVM, RandomForest, LogisticRegression, GradientBoosted, DecisionTrees. In addition, we also use Deep Learning models such as ROBERTA, DistilBERT, XLNet, Electra, BERT.\\
\\
These models were trained and evaluated using the Dreaddit dataset of Reddit posts.
\subsubsection{Training datasets:}
We use the Dreaddit dataset to train the model, which was divided into train and test sets with a ratio of 8:2, with the proportion of posts being quite similar in each set.
\subsubsection{Data preprocessing:} Data preprocessing is extremely important for any social network data analysis system, because it directly affects the complexity of the data. Although social networks are considered a valuable data source, they are the most complex type of text data, with the diversity of links, hashtags, special symbols, emoticons, and many other factors. Because of these factors, we preprocess the data from Reddit through the steps mentioned in section ~\ref{4.1}
\subsubsection{Machine Learning models:} We use SVM, RandomForest, LogisticRegression, GradientBoosted, and DecisionTrees models from Spark MLlib. Using these models for analysis can provide flexibility and diversity in processing, helping us optimize the performance of analysis on the data being processed in the Spark MLlib environment.
\subsubsection{Deep Learning models:}
The main goal of our team is to implement Deep Learning models on the Apache Spark big data platform. This is a challenge because previous research works have not been able to do this, as Spark does not support libraries for Deep Learning models. To solve this problem, we decided to use SparkNLP, an open source library built on Apache Spark and Spark ML (MLlib) library, to support the development of Deep Learning models directly on Spark.
\subsection{Online processing system}
The online system is a pipeline that detects stress emotions in Reddit posts in real time. The three main tasks in this part are collecting online data from Reddit, streaming and processing data in real time, and analyzing the emotions of this data in real time.
\subsubsection{Collecting online data from Reddit in real time:}
In this step, we use Reddit API to collect posts from various domains on Reddit. Reddit API was used to retrieve data about Reddit posts that were created in real time. To connect to the API and retrieve data, our team used the Python library Praw to collect Reddit data.
\subsubsection{Streaming online data:}
After receiving online data from Reddit, we use Apache Kafka to store and continuously send data to Spark Structured Streaming. Spark Structured Streaming provides the ability to automatically process when new data is added, helping to maintain real-time information. This makes the online data processing process robust and flexible, especially in dealing with dynamic and continuous data streams from Reddit.
\subsubsection{Real-time stress detection:}
Spark Structured Streaming processes the data collected in real time, converts them into vectors to be input for the best model trained in the offline part, and then this model detects stress from the corresponding posts. Basically, the steps of processing and predicting in real time are explained as follows:

- In the processing process, Spark Structured Streaming retrieves the content of the post on Reddit and performs the necessary preprocessing steps. Then, the data is extracted to form features and converted into vector structures before being fed into the trained model.

- For the prediction part, Spark uses the best prediction model that was trained before to detect stress and then outputs the prediction results in real time. This helps to identify and track changes in the stress level of the Reddit community over time, providing important information about the trends and moods of the users in that community.
\section{Experiment and results}
\subsection{Model experimental results}
In this problem, we will focus on evaluating models using two key evaluation metrics: Accuracy and F1-macro. Both metrics bring distinct advantages. Accuracy is chosen for its simplicity, providing a clear understanding of the ratio of correct predictions across the entire dataset. Additionally, to ensure a more comprehensive evaluation of the models, F1-macro will be used. F1-macro is chosen because it is not dependent on the size of each class, helping reduce the impact of data imbalance. This approach allows us to assess the model's classification ability more fairly, especially in the presence of significant variations in the number of samples between classes.

After preprocessing and data normalization steps, we proceeded to experiment with machine learning (ML) models: Logistic Regression (LR), Decision Tree (DT), Random Forest (RF), Gradient-Boosted Trees (GBT), and Support Vector Machines (SVM) using Spark ML. The Word2Vec embedding was employed to embed the words into a vector space. To mitigate the randomness of experimental results, we utilized the Cross Validation technique with K-Fold set to 10 and recorded the results in table\textbf{\textcolor{red}{~\ref{modelResults}}}.\\

\begin{table}[!ht]
\centering
\resizebox{\linewidth}{!}{
\begin{tabular}{|c|c|c|c|}
  \hline
  \multirow{5}{*}{\textbf{Machine Learning}} & \textbf{Model} & \textbf{Accuracy} (\%) & \textbf{F1-score} \\
  \cline{2-4}
  & LogisticRegression & 70,07 & 69,93 \\
  \cline{2-4}
  & Support Vector Machines (SVM) & 66,62 & 65,31 \\
  \cline{2-4}
  & RandomForest & 64,34 & 66,05 \\
  \cline{2-4}
  & Gradient-Boosted Trees & 62,87 & 63,61 \\
  \cline{2-4}
  & Decision Trees & 61,13 & 61,56 \\
  \hline
  \multirow{5}{*}{\textbf{Deep Learning}}
  & XLNet & 80,65 & 80,43 \\ 
  \cline{2-4}
  & BERT & 78,35 & 78,28 \\
  \cline{2-4}
  & DistilBERT & 76,21 & 75,89 \\
  \cline{2-4}
  & Electra & 75,86 & 76,11 \\
  \cline{2-4}
  & RoBERTa & 52,97 & 53,11 \\
  \hline
\end{tabular}
}
\caption{Evaluation results of model performance on sample data}
\label{modelResults}
\end{table}
The results from the table indicate that the Support Vector Machines (SVM) model achieved the highest accuracy at 66.62\% and an F1-score of 65.3. Logistic Regression, with stable accuracy at 65.95\%, had an F1-score of 66.36, demonstrating good performance in identifying both classes. Random Forest performed relatively well with an accuracy of 64.34\% and an F1-score of 66.05. Gradient-Boosted Trees and Decision Trees both had lower performance compared to other models, with accuracy and F1-score ranging from 61 to 63. The results suggest that ML models did not exhibit significant differentiation. Therefore, our group proceeded to implement and run experiments on the dataset using deep learning (DL) models such as BERT, RoBERTa, DistilBERT, XLNet, and Electra. All five models were configured with 10 epochs, a maximum sequence length of 512 to capture the entire content of posts and comments, a batch size of 8, and a learning rate of 1e-04. Among them, the XLNet model yielded the highest results with an accuracy of 80.65\% and an F1-score of 80.43 due to its large complexity with 335 million parameters and pre-training data having similarities with our dataset. The cluster of BERT, DistilBERT, and Electra models produced similar results with accuracy ranging from 75 to 78\% and F1-Score from 76 to 78. RoBERTa, however, yielded the lowest result at 53\% accuracy and a 53.11 F1-Score, lower than all machine learning models. We observed that during model training, RoBERTa could only predict rows with label=1 and failed to predict any rows with label=0, a phenomenon also observed with RoBERTa architecture models like XLM-RoBERTa, highlighting a potential issue in the training process.
\subsection{Experiment results based on realistic annotation}
We have collaborated with three students assigned the task of labeling 100 data samples collected directly from Reddit independently. After all three have completed labeling, the labeling results of the three individuals will use the majority label selection method to determine the final label for each data sample. This final label is considered the ground truth for the 100 data samples. Subsequently, we proceed with model predictions on these 100 data samples and compare them with the ground truth. This allows us to assess the real-world performance of the model with the latest directly updated data samples from Reddit.
\\
\begin{table}
\centering
\resizebox{\linewidth}{!}{
\begin{tabular}{|c|c|c|c|c|}
\hline
\textbf{Model} & \textbf{F1-score} & \textbf{Accuracy (\%)} & \textbf{Stress} & \textbf{Non-stress} \\
\hline
LogisticRegression & 68,97 & 69,39 & - & -\\
\hline
Support Vector Machines (SVM) & 66,09 & 62,24 & - & - \\
\hline
RandomForest & 62,59 & 64,29 & - & -\\
\hline
Gradient-Boosted Trees & 65,32 & 66,33 & - & -\\
\hline
Decision Trees & 61,39 & 62,24 & - & - \\
\hline
XLNet & 67,31 & 69, 11 & 75,08 & 59,77\\
\hline
BERT & 63,04 & 67,14 & 75,05 & 51,22 \\
\hline
DistilBERT & 60,38 & 64,09 & 72,12 & 47,01 \\
\hline
Electra & 64,67 & 68,11 & 75,33 & 53,96 \\
\hline
RoBERTa & 35,84 & 54,42 & 70,55 & 0 \\
\hline
\end{tabular}
}
\caption{Evaluation results of model when predicting new streaming data}
\label{tab2}
\end{table}

From the results of table\textbf{\textcolor{red}{~\ref{tab2}}}, we observe that the LogisticRegression machine learning model still produces the best results with an accuracy of 69.39\% and an F1-score of 68.97. The other machine learning models have lower accuracy and F1-score, but the differences are not significant. Accuracy ranges from 62 to 66\%, and F1-score ranges from 61 to 66\%. Notably, in this section, deep learning models have lower results than machine learning models. This could be attributed to the relatively small training data compared to the requirements of machine learning models, as well as the limited number of test data samples, causing deep learning models to not achieve high results. The highest-performing model is still XLNet with an accuracy of 67.31\% and an F1-score of 69.11. The remaining models have accuracy ranging from 60 to 64\% and F1-score from 64 to 69. RoBERTa, in particular, has the lowest accuracy and F1-score at 54.42\% and 35.84, respectively. Additionally, deep learning models demonstrate good recognition of samples labeled as stress, achieving relatively high F1-scores around 75, such as XLNet, BERT, and Electra. However, in recognizing samples labeled as non-stress, the F1-scores are considerably lower, ranging from 47 to 59, with DistilBERT being the lowest.
\section{Conclusions}
In this work, we have successfully deployed models using Machine Learning algorithms supported by Machine Learning Library (MLlib) as well as Deep Learning models. The results showed that the XLNet model achieved the best results, with F1-score of 80.43\% and accuracy of 80.65\% on the test set.\\

As this work focus on the practicality, we used Apache Kafka, Apache Spark combined with Apache Airflow to ensure the ability to process large amounts of data of the system. In addition, we also succeeded in building a system to detect stress in real time, combining big data technologies and integrating them into a complete system, focusing on training models with large amounts of data and applying the trained models to process the posts quickly and continuously.


\end{document}